\documentclass[10pt,twocolumn,letterpaper]{article}

\usepackage{iccv}
\usepackage{times}
\usepackage{epsfig}
\usepackage{graphicx}
\usepackage{amsmath}
\usepackage{amssymb}
\usepackage{multirow}
\usepackage{array}

\makeatletter
\newcommand{\thickhline}{%
    \noalign {\ifnum 0=`}\fi \hrule height 1pt
    \futurelet \reserved@a \@xhline
}
\newcolumntype{"}{@{\vrule width 1pt}}
\makeatother

\usepackage[pagebackref=true,breaklinks=true,letterpaper=true,colorlinks,bookmarks=false]{hyperref}

\iccvfinalcopy 

\ificcvfinal\fi

\begin{document}

\title{Spatiotemporal Transformer for Video-based Person Re-identification}

\author{
    Tianyu Zhang,\textsuperscript{\rm 1} Longhui Wei,\textsuperscript{\rm 5} Lingxi Xie,\textsuperscript{\rm 4} Zijie Zhuang,\textsuperscript{\rm 4} Yongfei Zhang,\textsuperscript{\rm 1,2,3} Bo Li,\textsuperscript{\rm 1,2,3} Qi Tian \textsuperscript{\rm 6}\\
    \small
    \textsuperscript{\rm 1}Beijing Key Laboratory of Digital Media, School of Computer Science and Engineering, Beihang University, \\
    \small
    \textsuperscript{\rm 2}State Key Laboratory of Virtual Reality Technology and Systems, Beihang University,\\
    \small
    \textsuperscript{\rm 3}Pengcheng Laboratory,
    \textsuperscript{\rm 4}Tsinghua University,
    \small
    \textsuperscript{\rm 5}University of Science and Technology of China,
    \textsuperscript{\rm 6}Xidian University\\
    \small
    \{tianyu1949, weilh2568, 198808xc, jayzhuang42, wywqtian\}@gmail.com, \{yfzhang, boli\}@buaa.edu.cn
}

\maketitle
\ificcvfinal\thispagestyle{empty}\fi

\begin{abstract}
Recently, the Transformer module has been transplanted from natural language processing to computer vision. This paper applies the Transformer to video-based person re-identification, where the key issue is to extract the discriminative information from a tracklet. We show that, despite the strong learning ability, the vanilla Transformer suffers from an increased risk of over-fitting, arguably due to a large number of attention parameters and insufficient training data. To solve this problem, we propose a novel pipeline where the model is pre-trained on a set of synthesized video data and then transferred to the downstream domains with the perception-constrained Spatiotemporal Transformer (STT) module and Global Transformer (GT) module. The derived algorithm achieves significant accuracy gain on three popular video-based person re-identification benchmarks, MARS, DukeMTMC-VideoReID, and LS-VID, especially when the training and testing data are from different domains. More importantly, our research sheds light on the application of the Transformer on highly-structured visual data.

\end{abstract}

\section{Introduction}

With increasing concerns on public security, person re-identification (ReID), which aims to re-identify certain pedestrians in a disjoint camera network, has drawn more and more attention from researchers. In past decades, image-based ReID tasks, which rely on stationary pedestrian appearance from isolated and noncontinuous frames, have been studied widely, and lots of novel works~\cite{CBN,SCT,PTGAN,GLAD,GLTR,JVTC} have been proposed.

However, image-based ReID tasks neglect several vital clues.
Compared to isolated frames, continuous frames from surveillance videos provide rich spatiotemporal information that reveals the causality between motions/viewing-angle-changes and identity-appearance-variations. 
Aggregating such spatiotemporal information properly into ReID tasks benefits deep ReID models in extracting discriminative and stable features against the misalignment of both body parts and pedestrian locations inside each frame.
Meanwhile, the redundancy of pedestrian appearance in continuous frames helps to overcome temporary occlusion naturally.
Thus, video-based ReID tasks, more precisely tracklet-based ReID tasks, have recently become more attractive for the ReID community~\cite{MGH,STGCN,AGRL,STA}.

\begin{figure}
    \centering
    \includegraphics[width=0.47\textwidth]{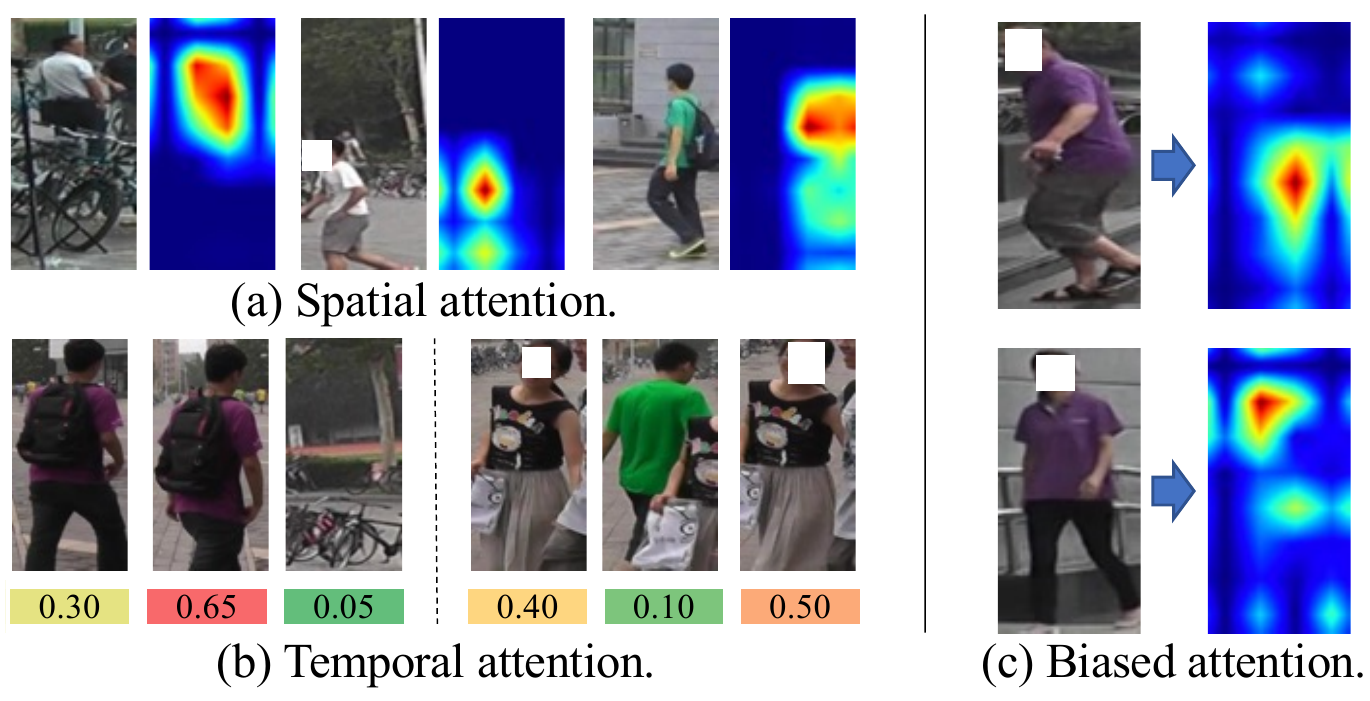}
    \label{fig:intro}
    \caption[]{Illustrations of the strengths and weaknesses of the Transformer. (a) Spatial attention is suitable for misaligned bounding boxes or occluded persons. (b) Temporal attention is good at excluding misdetection. (c) The vanilla Transformer learns biased attention. Minimal regions of the images are involved in the attention map.}
\end{figure}

The key for video-based ReID tasks is to efficiently aggregate the spatial and temporal knowledge provided by tracklets into one representation learning framework.
The Transformer~\cite{aayn} originated from natural language processing (NLP) tasks sheds light on long sequence modeling.
Encouraged by its success in NLP tasks~\cite{BERT,GPT3}, there are several attempts to utilize the Transformer in computer vision tasks~\cite{DETR,TTSR}.
Most of them focus on static and noncontinuous images.
Images are manually cropped as patches, and Transformers are used to model the relations of these patches.
Benefited from a strong capability of ``attention and then aggregation'', Transformer-based networks learn discriminative representations from all image patches and thus achieve notable success.

Such a capability meets the demands of video-based ReID tasks perfectly, where the learning procedure is required in not only the spatial dimension of each frame, {\em e.g.}, relations between human body parts but also the temporal dimension across continuous frames, {\em e.g.}, appearance variations through time.
Thus, we propose a two-stage \textbf{S}patio\textbf{t}emporal \textbf{T}ransformer (STT) module, where a Spatial Transformer handles the information in image patches and another subsequent Temporal Transformer focuses on the video sequence. 
As shown in Fig.~\ref{fig:intro}(a) and Fig.~\ref{fig:intro}(b), the Spatial Transformer is capable of attending to human regions from various backgrounds, while the Temporal Transformer excludes ``noisy'' frames in the given tracklet.
However, such a vanilla assembly of two Transformer modules suffers from the over-fitting risk due to the insufficient training data in video-based ReID tasks.
Fig.~\ref{fig:intro}(c) demonstrates a typical failure case.
Although the Transformer surpasses convolutional neural networks (CNN) in excluding background areas, it also tends to over-focusing on local salient body parts while ignoring other common but useful body parts information.
Without any restrictions on learning generic pedestrian appearance knowledge, the performance of the vanilla Transformer on video-based ReID tasks is much worse than previous CNN-based methods, ~\emph{e.g.}, $11.8\%$ rank-1 accuracy drop on MARS dataset.

To address the over-fitting issue mentioned above, we further propose the constrained attention learning scheme upon the STT module, where multiple prior constraints are applied on the Spatial Transformer and Temporal Transformer, respectively.
Meanwhile, we introduce a global attention learning branch as the supplement to exploit the relationships between patches from different frames. 
Finally, since the insufficient training data hinders us from fully validating the Transformer in video ReID, we resort to the synthesized data and propose a novel pre-training pipeline that relieves the over-fitting problem.
Extensive experiments demonstrate that, with our proposed pipeline, on three popular benchmarks,~\emph{i.e.}, MARS, DukeMTMC-VideoReID, and LS-VID, our STT outperforms both CNN baselines and the vanilla Transformer by a large margin.
Our contributions are summarized as follows:
\begin{itemize}
    \item We propose a simple yet effective framework with the Spatiotemporal Transformer for video-based person ReID. To the best of our knowledge, it is the first to evaluate the effectiveness of the Transformer in video-based ReID tasks, which paves a new way for further study.
    \item 
    To reduce the over-fitting risk caused by the Transformer and further enhance the diversity of learned representation, we propose a constrained attention learning scheme upon the STT module and an extra Global Transformer module as a supplement. Moreover, the synthesized video data are generated for pre-training.
    \item Extensive experiments on two large-scale video-based ReID datasets have validated the effectiveness of the proposed approaches. Compared with the previous state-of-the-art methods with pure CNN architectures, our Transformer-based approach significantly improves the final performance in multiple datasets.
\end{itemize}

\section{Related Work}
\subsection{Person Re-identification}
Existing image-based ReID methods mainly focus on discriminative feature extraction from images.
Human body parts are essential for image-based ReID.
Many study how to learn feature embeddings from detected body parts~\cite{GLAD,PSE}, or fixed image areas~\cite{MGN,PCB}.
Some works learn the relationship between body parts.
For example, Su~\emph{et al.}~\cite{PDC} align image patches that contain the same body part;
Xia~\emph{et al.}~\cite{SONA} utilize Non-local module to explore the importance of different parts;
He~\emph{et al.}~\cite{transreid} adopt the Vision Transformer~\cite{ViT} for image-based ReID feature extraction.
These methods show excellent performance on public datasets, such as Market-1501~\cite{MARKET} and MSMT17~\cite{PTGAN}.
However, the temporal information from video tracklets also provides helpful information to re-identify a person.
Therefore, researchers pay more and more attention to video-based ReID methods recently.
Early works introduce the optical flow and design two-stream networks to process optical flow and RGB frames, respectively.
Due to the computational complexity of the optical flow, many works turn to more direct ways of modeling image sequences, such as RNN~\cite{RNN1,RNN2,RNN3} and LSTM~\cite{videolstm,Yan2016PersonRV}.
Graph neural networks are also effective to mine the relationship of continuous frames~\cite{STGCN,MGH}.
Although these methods achieve promising performance, the extra heavy modules mentioned above are time-consuming.
Compared to these methods, the Transformer enjoys the benefits of time-efficiency and is more suitable for video-based ReID tasks.

\begin{figure*}
    \centering
    \includegraphics[width=0.97\textwidth]{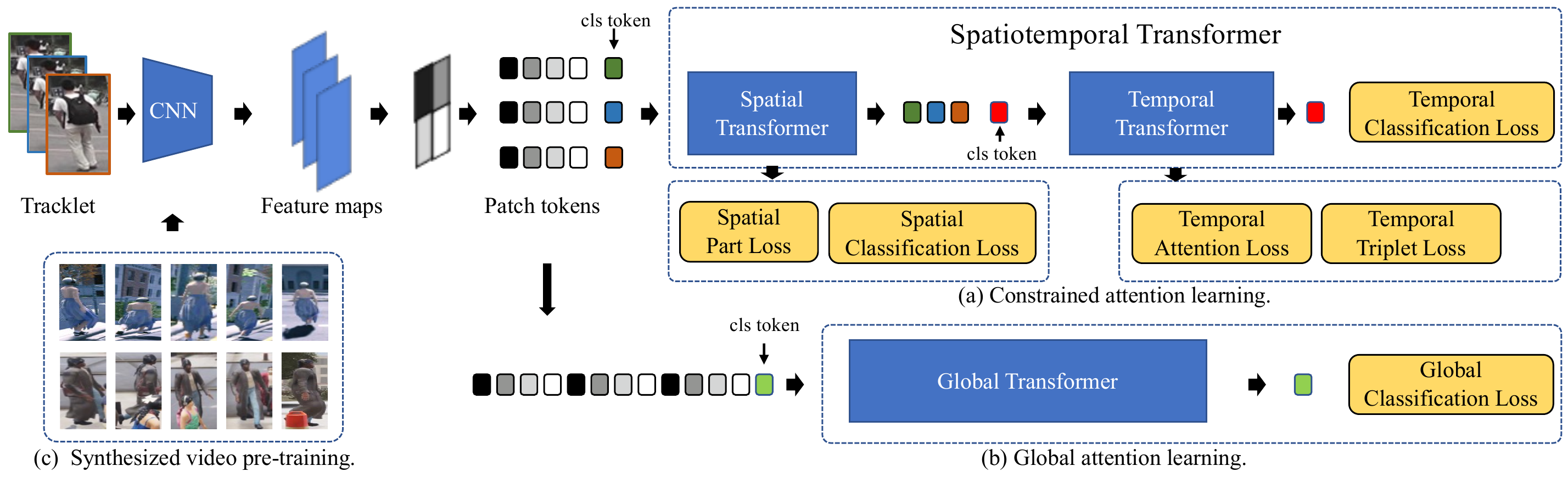}

    \caption{The overall framework with the Spatiotemporal Transformer. (a) The constrained attention learning is adopted on the Spatial Transformer and Temporal Transformer. (b) A global attention learning branch is proposed to supplement STT. (c) The synthesized video pre-training plays an important role in avoiding over-fitting. }
    \label{fig:framework}
\end{figure*}

\subsection{Transformer for Vision Tasks}

Transformer~\cite{aayn} is a self-attention-based architecture proposed for modelling sequence data.
In natural language processing (NLP), it has become the basic module for state-of-the-art methods~\cite{BERT,GPT3}.
Attention mechanism has also been studied in computer vision and plays an important role in many computer vision tasks.
More recently, researchers apply the Transformer directly on image patches, like Vit~\cite{ViT} and DeiT~\cite{DeiT}.
With large-scale pre-training~\cite{ViT}, or distillation techniques~\cite{DeiT}, these methods achieve competitive results on image classification tasks.
The application of Transformer modules on video tasks has also arisen.
For example, Gavrilyuk~\emph{et al.}~\cite{actortrans} propose the actor-transformer for activity recognition.
To better encode the point cloud for the 3D video object detection task, Yin~\emph{et al.}~\cite{lidar3d} propose a Spatial Transformer to distinguish background and a Temporal Transformer to integrate motion information.
In this paper, we propose a hybrid architecture of CNN and Transformer encoders for extracting features from each frame and aggregating appearance information from the whole tracklet.
To the best of our knowledge, this is the first work that applies the Transformer to video-based person ReID.

\section{Methods}
\subsection{Formulation}

Given an annotated video dataset $\mathcal{S}=\{(\boldsymbol{T}_{1}, \boldsymbol{y}_{1}), (\boldsymbol{T}_{2}, \boldsymbol{y}_{2}), ..., (\boldsymbol{T}_{N}, \boldsymbol{y}_{N})\}$, where each $\boldsymbol{T}_i$ denote a tracklet, and $\boldsymbol{y}_i$ is the ground truth label of the corresponding identity, the goal of video-based ReID tasks is to learn a feature embedding function $\boldsymbol{f}(\theta;\boldsymbol{T}_i)$ that maps tracklets into a feature space $\mathcal{X}=\{\boldsymbol{x}_i|\boldsymbol{x}_i=\boldsymbol{f}(\theta;\boldsymbol{T}_i),1\leq i \leq N\}$, where tracklets of the same identity are closer than that of different identity. 
To achieve this goal, researchers propose two types of objective functions, {\em i.e.}, classification loss and distance metric learning loss.
The classification loss classifies features $\boldsymbol{x}_i$ of the same person into the same category, which can be implemented by the cross entropy loss:
\begin{equation}\label{eq:xent}
    \mathcal{L}_\mathrm{xent}=\sum^{N}_{i=1} -\log (\frac{\exp(\boldsymbol{h}(\boldsymbol{x}_i)^c)}{\sum_{j}\exp(\boldsymbol{h}(\boldsymbol{x}_i)^j)}),
\end{equation}
where $\boldsymbol{h}$ is a classifier, consisting of a linear layer and softmax layer that maps $\boldsymbol{x}$ into a one-hot classification vector.
For distance metric learning losses, the triplet loss with online hard sample mining~\cite{InDefense} is a popular choice.
For each mini-batch, it demands each positive sample pair closer than all negative samples by at least a margin:

\begin{equation}\label{eq:trip}
\mathcal{L}_\mathrm{trip}= \sum_{i=1}^C [m+ \mathrm{dist}(\boldsymbol{x}_i,p(\boldsymbol{x}_i))  - \mathrm{dist}(\boldsymbol{x}_i,n(\boldsymbol{x}_i))]_+,
\end{equation}
where $[z]_+=\max(z,0)$, $m$ denotes the margin.
$p(\boldsymbol{x}_i)$ is the farthest sample belonging to the same person as $\boldsymbol{x}_i$, and $n(\boldsymbol{x}_i)$ is another identity's image that is the nearest to $\boldsymbol{x}_i$.

In video-based ReID tasks, each tracklet consists of a series of images, which are usually generated by tracking techniques,~\emph{i.e.}, $\boldsymbol{T}_{i}=\{\boldsymbol{I}_{i}^{(1)},\boldsymbol{I}_{i}^{(2)},...,\boldsymbol{I}_{i}^{(n)}\}$.
Compared to image-based ReID, video-based ReID tasks introduce a new temporal dimension alongside existing spatial dimensions.
Thus, the major challenges in video-based ReID tasks are usually two-fold.
First, a good feature embedding for each image is required to encode appearance characteristics and spatial information.
Second, which is also the main difference compared to image-based ReID tasks, video-based ReID tasks demand an ingenious module for modeling temporal relations among all frames in each tracklet.

However, modeling temporal relations is extremely challenging in real-world applications, especially for a non-ideal, unconstrained surveillance video.
Both the severe mismatch of human body parts and foreground position shifts inside each frame might lead to significant performance decreases.
To tackle these issues, we disassemble the mismatch problem into two categories: spatial mismatch in the image-level and temporal mismatch in the tracklet-level, and attempts to adopt Transformer-based methods to address the above challenges accordingly. 
Details of our method are presented in the next section.

\subsection{Overall Framework}

Benefited from the strong capability of ``attention and aggregation'', we can directly utilize Transformers for video-based ReID tasks, extracting and aggregating the useful human information spatially and temporally while neglecting disturbances such as occlusions and background areas.

The overall of our proposed framework is shown in Fig.~\ref{fig:framework}. 
We first utilize a truncated convolutional neural network, {\em e.g.}, a ResNet-50 backbone with the first three blocks, as the preliminary feature encoder.
For each frame, the CNN backbone encodes it into a 3-dimensional feature map.
Next, each feature map is split into small patches.
We regard each patch as a ``token'' and feed tokens from the same image into our Spatial Transformer (ST) module. 
In this way, the ST module learns the spatial relations among all given tokens and aggregates discriminative spatial information as well as appearance features from them.
Simultaneously, following the practice in~\cite{ViT,DeiT}, we add another trainable classification token for fusing the spatial information of each input image.
This token fuses the information of all patch-based tokens.
Since it further encodes the spatial relation between tokens, we name it the spatial token.
For each frame, this spatial token is regarded as the final output of the ST module.
After the ST module encodes each frame, the Temporal Transformer (TT) module collects all spatial tokens from the same tracklet and aggregates them into a temporal token.
The final temporal token embeds the discriminative identity information spatially and temporally.
It is used as the final representation for the entire tracklet.

The vanilla two-stage Transformer mentioned above processes both spatial and temporal information at the same time, making it conceptually suitable for video-based ReID tasks.
However, when using it in video-based ReID tasks directly, it suffers from severe over-fitting.
As shown in Tab.~\ref{tab:abl1} and Tab.~\ref{tab:abl2}, compared to the CNN baseline, this framework reports much worse performance, even though its preliminary feature encoder is the same as the first three blocks of the CNN baseline.
We attribute this performance decrease to three reasons.
First, the Transformer is heavily parameterized.
Without any restrictions, it could easily fall into the local minimal and generalize poorly on the testing data.
To address this issue, we propose a constrained attention learning scheme especially to prevent the Transformer from over-focusing on local regions.
Second, the two-stage design separates the spatial information and temporary information to avoid the over-fitting.
However, with this design, patches from one image cannot communicate with any other patches of another image.
Thus, we add a global attention learning branch that associates patches across frames.
Third, since the Transformer introduces much more parameters, existing video-based ReID datasets are insufficient to tune these parameters fully.
To tackle this problem, we introduce pre-trained video-based ReID models with synthesized images to achieve better initialization.
In the following section, we introduce our design principles and the proposed pipeline in detail.

\begin{figure}
    \centering
    \includegraphics[width=0.47\textwidth]{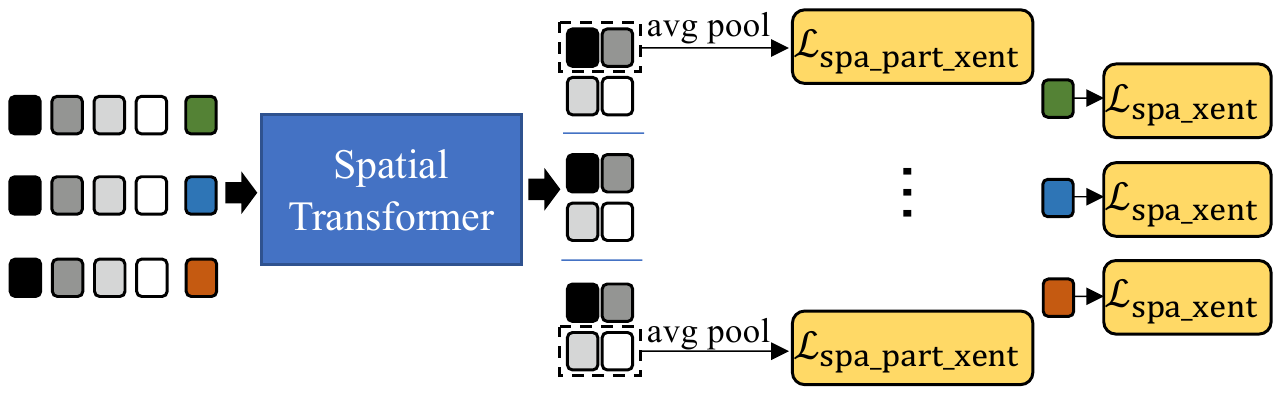}
    \caption{Illustration of the spatial constraint. The tokens of patches are reshaped for horizontal average pooling.}
    \label{fig:spac}
\end{figure}

\subsection{Constrained Attention Learning}

Owing to the two-stage architecture of the Spatiotemporal Transformer, which contains the Spatial Transformer (ST) to process the patch tokens in image-level and the Temporal Transformer (TT) to handle the image tokens in tracklet-level, we set different constraints on ST and TT individually to relieve the over-fitting problem.

\subsubsection{Spatial Constraint}
This Spatial Transformer module of STT processes patch tokens in each image.
To be more specific, each feature map extracted by the CNN backbone is split into $H\times W$ patches and then flattened into $1$-dim tokens.
Therefore, with the extra classification token (or called spatial token in this paper), the ST module receives $H\times W + 1$ tokens in total.
Aiming to learn discriminative representation in spatial dimension, we add the cross entropy loss, ~\emph{i.e.}, $\mathcal{L}_\mathrm{spa\_xent}$, on the spatial token.
Therefore, the spatial token is forced to focus on the human information to classify the input.
Because of the small-scale data for video ReID, ST can easily focus on limited regions but ignore detailed cues (~\emph{e.g.}, backpack). To further avoid this, we add a spatial part cross entropy loss,~\emph{i.e.}, $\mathcal{L}_\mathrm{spa\_part\_xent}$, to force each token to learn useful recognition information as much as possible. The spatial part loss is motivated by the recent part-based image ReID methods~\cite{GLAD,PCB}.
Specifically, as shown in Fig.~\ref{fig:spac}, we divide the $H\times W$ tokens into $P$ groups horizontally so that each group has $H/P \times W$ tokens.
An average pooling operation is conducted within every group.
Finally, a cross entropy classification loss is calculated for every averaged token, and the sum is formulated as the spatial part cross entropy loss:
\begin{equation}
    \mathcal{L}_\mathrm{spa\_part\_xent} = \frac{1}{P} \sum_{p=1}^{P} \mathcal{L}_\mathrm{spa\_part\_xent}^{(p)},
\end{equation}
where $\mathcal{L}_\mathrm{spa\_part\_xent}^{(p)}$ represents the cross entropy loss for the $p\mathrm{th}$ horizontal group. Obviously, this loss term leads the model to discover person-related clues in every horizontal part of the body, which makes sure each patch token learns enough human information. 

Therefore, the spatial constraint loss can be summarized as follows:

\begin{equation}
    \mathcal{L}_\mathrm{SpaC} = \mathcal{L}_\mathrm{spa\_part\_xent} + \mathcal{L}_\mathrm{spa\_xent}.
\end{equation}

\begin{figure}
    \centering
    \includegraphics[width=0.47\textwidth]{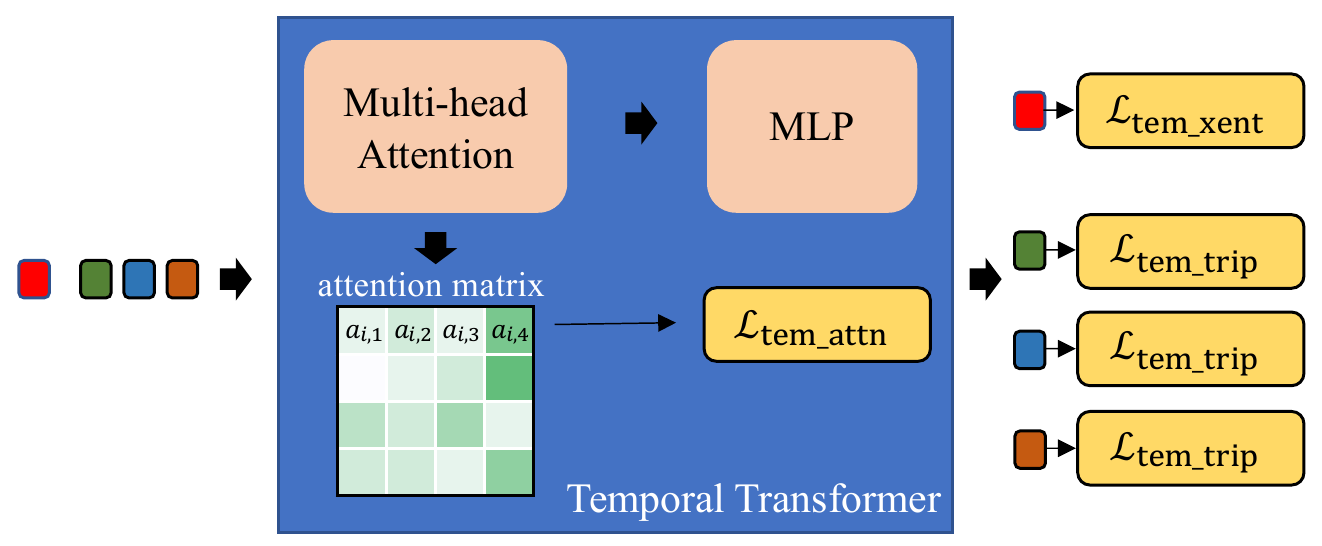}
    \caption{Illustration of the temporal constraint. Normalization layers and skip connections are omitted in this figure for clear presentation. The first row of the attention matrix (\emph{i.e.}, weights of classification tokens) is taken as the input of the temporal attention loss.}
    \label{fig:temc}
\end{figure}

\subsubsection{Temporal Constraint}
As shown in Fig.~\ref{fig:temc}, the Temporal Transformer (TT) module takes the spatial tokens from the Spatial Transformer as input.
One extra classification token (or called the temporal token) is added as the final feature representation.
In a tracklet sequence, the pedestrian may be occluded in some images but occur again in other images.
With the cross entropy loss $\mathcal{L}_\mathrm{tem\_xent}$ that supervises the final output, we expect the Temporal Transformer to exclude frames of lousy quality and utilize more information from frames that contain complete human bodies.
However, in the meantime, the risk of over-fitting is raised because the learning algorithm is not aware of the targeted human body.
For instance, an image containing a unique distractor person is also quite discriminative for the model.
Therefore, the Transformer module will attend to this image but ignore the targeted human.
More examples can be seen in Fig.~\ref{fig:vis_tem}.
To solve this problem, we expect the exploited representation should contain the common information of all frames in the tracklet and other specific information on each frame. 

To achieve this, apart from the cross entropy loss that supervises the final output, we add supervision for the temporal output tokens of each frame, named $\mathcal{L}_\mathrm{tem\_trip}$.
As shown in Eq.~\ref{eq:trip}, the triplet loss shrinks the distances of positive pairs. 
Because frames within the same tracklet are assigned with the same identity label, their distances are naturally reduced so that the extracted information is shared by most of the frames.
Besides, we also encourage the Temporal Transformer to focus on as many frames as possible to avoid biased attention to specific frames.
We add a temporal attention loss on the attention weights of the final classification token:
\begin{equation}
     \mathcal{L}_{\mathrm{tem\_attn}} = \sum_{i=1}^{N} [\exp(\sum_{k=1}^{L} \boldsymbol{a}_{i,k}\log(\boldsymbol{a}_{i,k})) - \alpha]_{+},
\end{equation}
where $L$ is the number of frames within a tracklet, $\boldsymbol{a}_{i,k}$ is the attention weight of the $k\mathrm{th}$ frame in the $i\mathrm{th}$ tracklet, and $\alpha$ is a hyper-parameter that adjusts the upper bound.
The temporal attention loss increases the information entropy of the attention weights in each tracklet.
Also, it leaves much space for the Transformer to decide which frame is more critical with the parameter $\alpha$.

Finally, the temporal constraint loss is as follows:
\begin{equation}
    \mathcal{L}_\mathrm{TemC} = \mathcal{L}_\mathrm{tem\_trip} + \mathcal{L}_\mathrm{tem\_attn}.
\end{equation}

\subsection{Global Attention Learning}
As described above, the Spatiotemporal Transformer learns image-level attention and tracklet-level attention in succession.
Therefore, the relationships between patches of different frames are ignored.
To address this problem, we further introduce the global attention learning branch in our framework.
This part has a Global Transformer module, which takes the feature map patches of all frames in a tracklet as the input.
More specifically, supposing we have $H\times W$ patches for every frame and $L$ frames in a tracklet, we concatenate these tokens of every frame within the same tracklet, resulting in an input, of which the length is $H \times W \times L$.
These are directly fed into the Global Transformer (GT) with an extra classification token. Then,
a cross entropy loss $\mathcal{L}_\mathrm{global\_xent}$ is adopted to supervise the learning of GT.
When the global attention learning branch is enabled, the final representation is generated by the concatenation of STT outputs and GT outputs.

\begin{table*}[htbp]
    \centering

    \resizebox{0.97\textwidth}{!}{
        \setlength{\tabcolsep}{0.12cm}
        \begin{tabular}{l|c|c|c|c|cc|cc|cc}
            \thickhline
            \multicolumn{1}{c|}{\multirow{2}[0]{*}{Models}} & \multirow{2}[0]{*}{Spatial Constraint} & \multirow{2}[0]{*}{Temporal Constraint} & \multirow{2}[0]{*}{Global Attention} & \multirow{2}[0]{*}{Pre-training} & \multicolumn{2}{c|}{MARS} & \multicolumn{2}{c|}{Duke} & \multicolumn{2}{c}{LS-VID} \\
            \cline{6-11}
            &       &       &       &       & rank-1 & mAP   & rank-1 & mAP   & rank-1 & mAP \\
            \hline
            CNN baseline &       &       &       &       & 84.1  & 78.8  & 50.6  & 48.3  & 16.6  & 11.2 \\
            \hline
            CNN+Transformer* &       &       &       &       & 72.3  & 63.1  & 24.5  & 24.5  & 4.5   & 3.9 \\
            CNN+Transformer & \checkmark &       &       &       & 87.1  & 83.5  & 60.5  & 56.3  & 16.2  & 11.7 \\
            CNN+Transformer & \checkmark & \checkmark &       &       & 87.2  & 84.2  & 61.1  & 57.4  & 18.2  & 13.1 \\
            CNN+Transformer & \checkmark & \checkmark &       & \checkmark & 88.6  & 83.9  & 62.4  & 61.0  & 18.7  & 13.1 \\
            \hline
            CNN+Transformer & \checkmark & \checkmark & \checkmark &       & 87.3  & 85.0  & 65.0  & 62.1  & 21.0    & 14.6 \\
            CNN+Transformer & \checkmark & \checkmark & \checkmark & \checkmark & \textbf{88.7} & \textbf{86.3} & \textbf{69.2} & \textbf{66.2} & \textbf{24.3} & \textbf{17.2} \\
            \thickhline
        \end{tabular}%
    }
    \caption{Evaluation results when training on MARS. * denotes the vanilla version of the Transformer.}
    \label{tab:abl2}%
\end{table*}%

\begin{table*}[htbp]
    \centering

    \resizebox{0.97\textwidth}{!}{
        \setlength{\tabcolsep}{0.12cm}
        \begin{tabular}{l|c|c|c|c|cc|cc|cc}
            \thickhline
            \multicolumn{1}{c|}{\multirow{2}[0]{*}{Models}} & \multirow{2}[0]{*}{Spatial Constraint} & \multirow{2}[0]{*}{Temporal Constraint} & \multirow{2}[0]{*}{Global Attention} & \multirow{2}[0]{*}{Pre-training} & \multicolumn{2}{c|}{MARS} & \multicolumn{2}{c|}{Duke} & \multicolumn{2}{c}{LS-VID} \\
            \cline{6-11}
            &       &       &       &       & rank-1 & mAP   & rank-1 & mAP   & rank-1 & mAP \\
            \hline
            CNN baseline &       &       &       &       & 34.1  & 18.8  & 91.2  & 88.0  & 20.8  & 13.7 \\
            \hline
            CNN+Transformer* &       &       &       &       & 23.2  & 12.8  & 90.7  & 88.2  & 7.3   & 5.9 \\
            CNN+Transformer & \checkmark &       &       &       & 45.0  & 28.2  & 96.2  & 96.0  & 32.0    & 22.0 \\
            CNN+Transformer & \checkmark & \checkmark &       &       & 46.4  & 28.9  & 96.6  & 96.0  & 32.9  & 22.6 \\
            CNN+Transformer & \checkmark & \checkmark &       & \checkmark & 51.4  & 33.6  & 97.4  & 97.0  & 34.6  & 24.3 \\
            \hline
            CNN+Transformer & \checkmark & \checkmark & \checkmark &       & 48.6  & 30.8  & 96.7  & 96.6  & 34.7  & 24.6 \\
            CNN+Transformer & \checkmark & \checkmark & \checkmark & \checkmark & \textbf{56.5} & \textbf{39.1} & \textbf{97.6} & \textbf{97.4} & \textbf{41.3}  & \textbf{29.4} \\

            \thickhline
        \end{tabular}%

    }

    \caption{Evaluation results when training on Duke. * denotes the vanilla version of the Transformer. }
    \label{tab:abl1}%

\end{table*}%

\begin{table*}[!t]
    \centering

    \resizebox{0.97\textwidth}{!}{
        \setlength{\tabcolsep}{0.12cm}
        \begin{tabular}{l|c|c|c|c|cc|cc|cc}
            \thickhline
            \multicolumn{1}{c|}{\multirow{2}[0]{*}{Models}} & \multirow{2}[0]{*}{Spatial Constraint} & \multirow{2}[0]{*}{Temporal Constraint} & \multirow{2}[0]{*}{Global Attention} & \multirow{2}[0]{*}{Pre-training} & \multicolumn{2}{c|}{MARS} & \multicolumn{2}{c|}{Duke} & \multicolumn{2}{c}{LS-VID} \\
            \cline{6-11}
            &       &       &       &       & rank-1 & mAP   & rank-1 & mAP   & rank-1 & mAP \\
            \hline
            CNN baseline &       &       &       &       & 36.1  & 20.3  & 50.6  & 43.2  & 58.5  & 44.6 \\
            \hline
            CNN+Transformer* &       &       &       &       & 12.0  & 7.7   & 16.8  & 17.9  & 31.9  & 28.3 \\
            CNN+Transformer & \checkmark &       &       &       & 50.0  & 32.7  & 74.2  & 71.3  & 78.8  & 68.0 \\
            CNN+Transformer & \checkmark & \checkmark &       &       & 52.7  & 35.4  & 77.8  & 74.9  & 80.7  & 70.5 \\
            CNN+Transformer & \checkmark & \checkmark &       & \checkmark & 54.0  & 37.8  & 83.1  & 80.5  & 84.6  & 74.7 \\
            \hline
            CNN+Transformer & \checkmark & \checkmark & \checkmark &       & 53.9  & 36.8  & 79.6  & 76.0  & 82.4  & 73.0 \\
            CNN+Transformer & \checkmark & \checkmark & \checkmark & \checkmark & \textbf{59.6} & \textbf{42.7} & \textbf{86.0} & \textbf{83.5} & \textbf{87.5} & \textbf{78.0} \\
            \thickhline
        \end{tabular}%
    }
    \caption{Evaluation results when training on LS-VID. * denotes the vanilla version of the Transformer.}
    \label{tab:abl3}%
\end{table*}%

\subsection{Synthesized Video Pre-training}
Previous works~\cite{ViT,DeiT} show that the data scale is one of the most important keys for the Transformer and conclude that Transformers can work better than pure CNN networks when the training data is sufficient.
Limited by the difficulty of annotations, it is nearly impossible to annotate a super large-scale video ReID dataset. 
To further relieve the over-fitting problem of the Transformer in the data aspect, we resort to the synthesized data and pre-train our model on this dataset.
Specifically, we adopt the UnrealPerson~\cite{UnrealPerson} toolkit to generate videos in 3D virtual scenes.
Four different environments and $34$ cameras are used in our synthesized data.
We implement two extra modifications to UnrealPerson.
First, we add large disturbance when cutting the bounding boxes.
In real scenes, the bounding boxes for video-based ReID tasks are not ideally aligned.
Therefore, persons in bounding boxes may not appear in the middle.
This helps the Transformer to concentrate on local patches where the body parts exist.
Second, to mimic the tracklets by tracking algorithms, we also keep the severely occluded frames.
Therefore, in a tracklet, the quality of images may vary largely.
This increases the difficulty of the synthesized data and encourages the model to find images of good quality in a tracklet.
The synthesized dataset we adopt in the experiments contains $90\mathrm{,}673$ tracklets of $2\mathrm{,}808$ identities.


\section{Experiments}

\subsection{Datasets}
We conduct experiments on two benchmark datasets, MARS~\cite{MARS}, DukeMTMC-VideoReID~\cite{dukevideo} and LS-VID~\cite{GLTR}.
MARS consists of $1\rm{,}261$ pedestrians captured under $6$ different cameras on campus.
The training set of MARS has $8\rm{,}298$ tracklets of $625$ persons, and the testing set has $621$ persons and $9\rm{,}330$ gallery images.
There are $1\rm{,}840$ valid query tracklets in total.
In MARS, each tracklet has 59 images on average.
DukeMTMC-VideoReID (referred as Duke for short) has $2\rm{,}196$ tracklets of $702$ pedestrians for training, and $2\rm{,}336$ tracklets of $1\rm{,}110$ persons for testing.
Each tracklet of Duke has $168$ frames on average.
LS-VID is constructed with videos from $15$ cameras. The training set contains $2\rm{,}831$ tracklets of $842$ identities, and the testing set includes $11\rm{,}333$ tracklets of $2\rm{,}730$ identities, among which $3\rm{,}504$ tracklets are used as queries.
Images and tracklets of MARS and LS-VID are generated automatically by detection and tracking algorithms,~\emph{i.e.}, DPM~\cite{DPM}, GMMCP~\cite{GMMCP} and Faster-RCNN~\cite{fasterrcnn}, while Duke is manually labeled.
In our experiments, we follow the standard evaluation protocols~\cite{MARKET} and report rank-1 accuracy and mean average precision (mAP).

\subsection{Implementation Details}
\textbf{Architecture. }We adopt the first three residual blocks of ResNet-50~\cite{resnet} that is pre-trained on ImageNet~\cite{ImageNet} as the CNN backbone.
The CNN baseline model has the complete layers of all $4$ residual blocks of ResNet-50, while in the proposed CNN+Transformer architecture, the $4\mathrm{th}$ residual block is replaced by Transformer blocks.
For our proposed Spatiotemporal Transformer, we follow the implementation of ViT~\cite{ViT}.
The output feature maps of the CNN backbone go through a convolutional layer and are flattened to patch tokens.
The Spatial and Temporal Transformers share the same architecture design, with $1$ layer and $6$ heads.
The Global Transformer has $2$ layers and $6$ heads.
The embedding dimension of all Transformers is set to $768$.
Positional embeddings are only used in the Spatial Transformer.
By discarding positional embeddings in the Temporal Transformer and Global Transformer, our framework is able to process tracklets of any length.

\textbf{Input.} Inspired by DeiT~\cite{DeiT}, we use more extreme data augmentation operations in our experiments, including random erasing~\cite{RandomErasing}, random patch~\cite{RPT1,RPT2}, random crop, random flip and AutoAugment~\cite{autoaug}.
The input size of images is set as $256\times 128$.
We sample $8$ images with the restricted sampling strategy~\cite{Li2018DiversityRS} from each tracklet when training.
For evaluation, we sample $32$ images from each tracklet.
A balanced sampling strategy~\cite{InDefense} is adopted to form the mini-batches. Specifically, $15$ identities are selected first, and for each identity, $4$ tracklets are chosen.
Thus the mini-batch size is $60$.
The size of the output feature map in the CNN backbone is $1024\times 16\times 8$, among which 1024 is the channel number.
The feature map is further embedded into $8\times 4$ patches by a convolution layer.

\textbf{Training Details. }For all cross entropy losses, we add label smoothing regularization with $\epsilon=0.1$.
The part number of the spatial part loss is $4$.
The margin of the temporal triplet loss is set as $0.3$.
In the temporal attention loss, the parameter $\alpha$ is set to $0.15$ empirically.
Besides loading the pre-trained weights for the CNN backbone, we also add an extra cross entropy loss for it to stabilize training.
In all our experiments, we adopt AdamW~\cite{Loshchilov2019DecoupledWD,ADAM} optimizer with the weight decay of $5e-4$.
The learning rate is set to $3.5e-4$ initially and multiply by a factor $0.1$ at the $80\mathrm{th}$ epoch and $160\mathrm{th}$ epoch, respectively.
The learning rate of all classifiers is set to a larger value, $1.75e-3$.
During the first $200$ iterations, the learning rate warm-up strategy is adopted.
All of the models are trained for $200$ epochs.

\subsection{Ablation Study}
In the ablation study, we evaluate the effectiveness of our proposed framework and the extra constraints for Transformer modules.
The results are shown in Tab.~\ref{tab:abl2}, Tab.~\ref{tab:abl1} and Tab.~\ref{tab:abl3}.
The direct transfer evaluation is also conducted among different datasets to validate the generalization ability of the trained models.

\textbf{Vanilla Transformer.} Compared to the pure CNN baseline model, the Transformer network without any additional constraints suffers performance drop, especially when trained on MARS and LS-VID.
The direct transfer performance decays even more than self-domain evaluation, indicating serious over-fitting exists in the trained model.
The reason has been illustrated in Fig.~\ref{fig:intro}(c),~\emph{i.e.}, an unconstrained Transformer network produces biased attention weights.

\textbf{Constrained Attention Learning.} When we add the spatial constraint losses in the model, the performance surpasses the pure CNN baseline by a large margin, especially on direct transfer results.
For example, when we evaluate Duke with the model trained on MARS, the rank-1 accuracy raises from $50.6\%$ to $60.5\%$.
The temporal constraint also improves all performances slightly.
These results demonstrate that our proposed loss terms are effective in preventing over-fitting.

\textbf{Global Attention Learning. }The global attention learning empowers each patch token to associate with patches from other frames.
Experiment results show that this global scope is an important complementary component for the Spatiotemporal Transformer.
For instance, in Tab.~\ref{tab:abl2}, when trained on MARS and evaluated on Duke, the model with global attention learning improves the rank-1 accuracy by $3.9\%$ compared to the model without global attention.

\textbf{Synthesized Data Pre-training. }The pre-training weights provide a good initialization for the Transformer modules, avoiding over-fitting from early training stages.
Note that we have the pre-trained weights fine-tuned on the downstream datasets for $200$ epochs, the same as training without synthesized data pre-training.
Therefore, there should leave very little information about the synthesized dataset in the final model weights.
Nevertheless, the direct transfer performance is boosted significantly on all three datasets, indicating that the Transformer benefits from better pre-trained weights and reaches a more optimal convergence in the end.

\subsection{Visualization Analysis}
\begin{figure}
    \centering
    \includegraphics[width=0.47\textwidth]{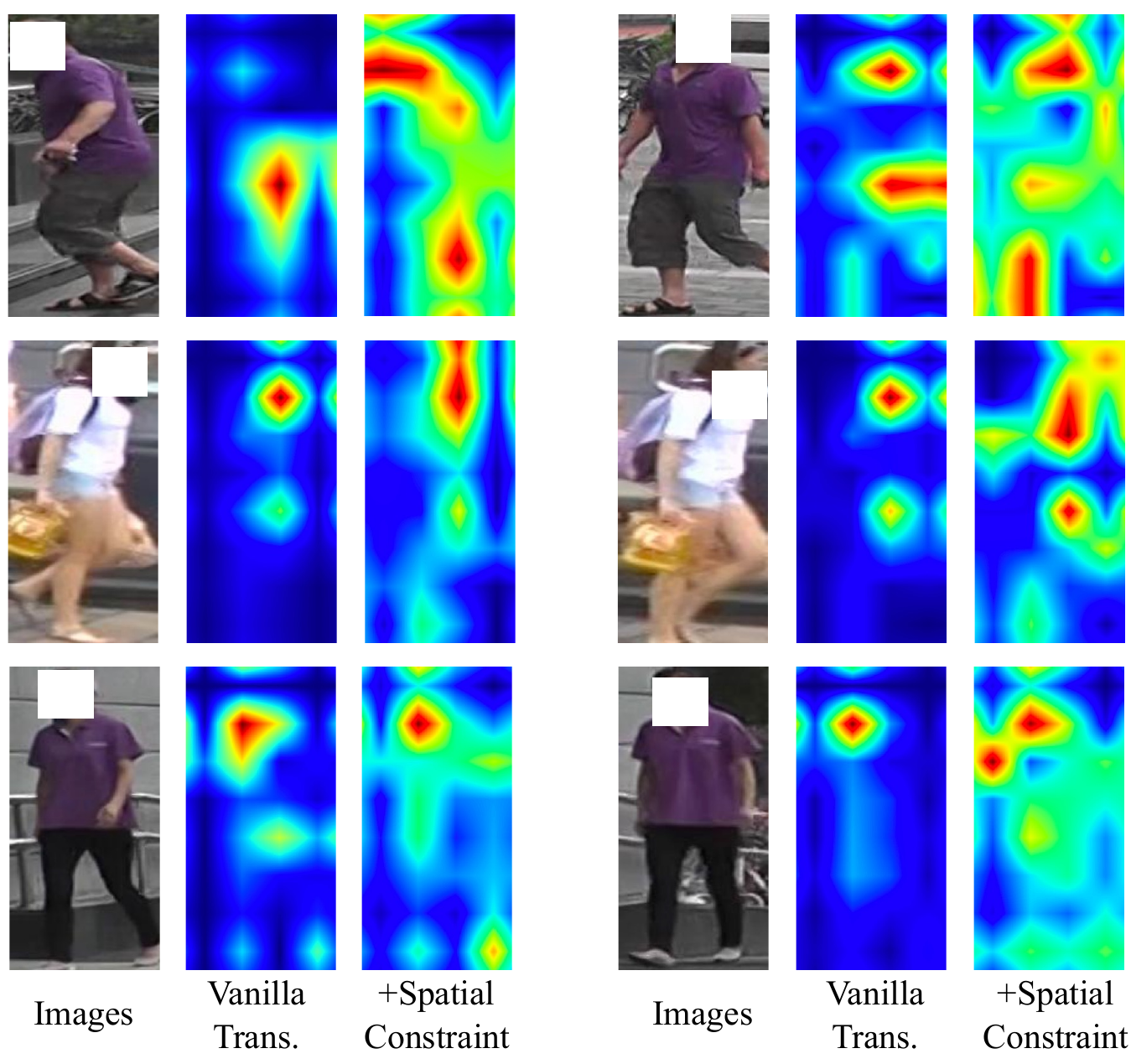}
    \caption{Spatial attention maps after adding spatial constraint losses.}
    \label{fig:vis_spa}
\end{figure}
\begin{figure*}
    \centering
    \includegraphics[width=0.97\textwidth]{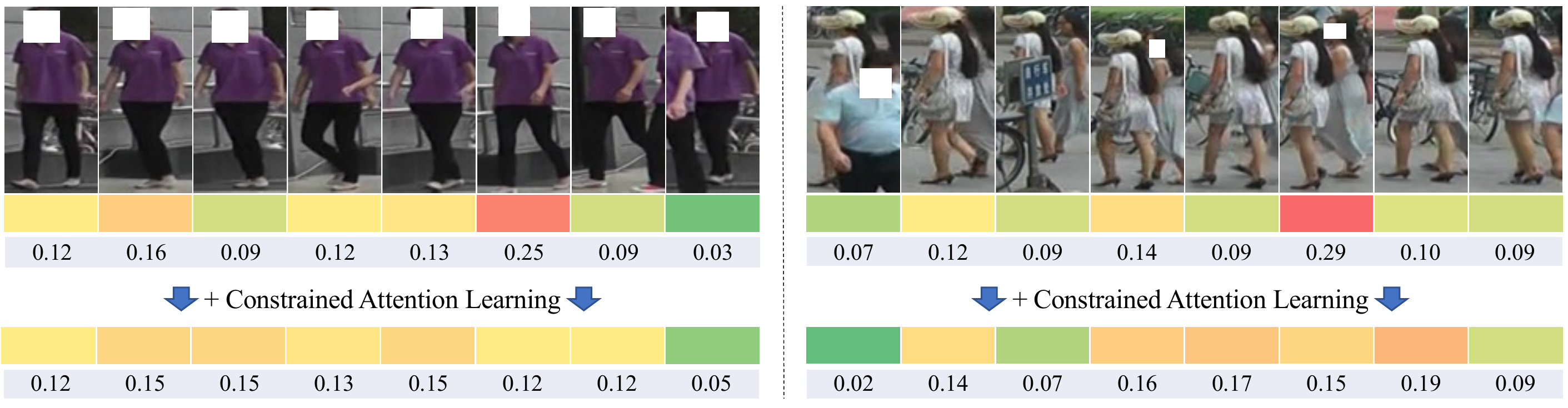}
    \caption{Temporal attention weights after using constrained attention learning.}
    \label{fig:vis_tem}
\end{figure*}
\begin{figure}
    \centering
    \includegraphics[width=0.47\textwidth]{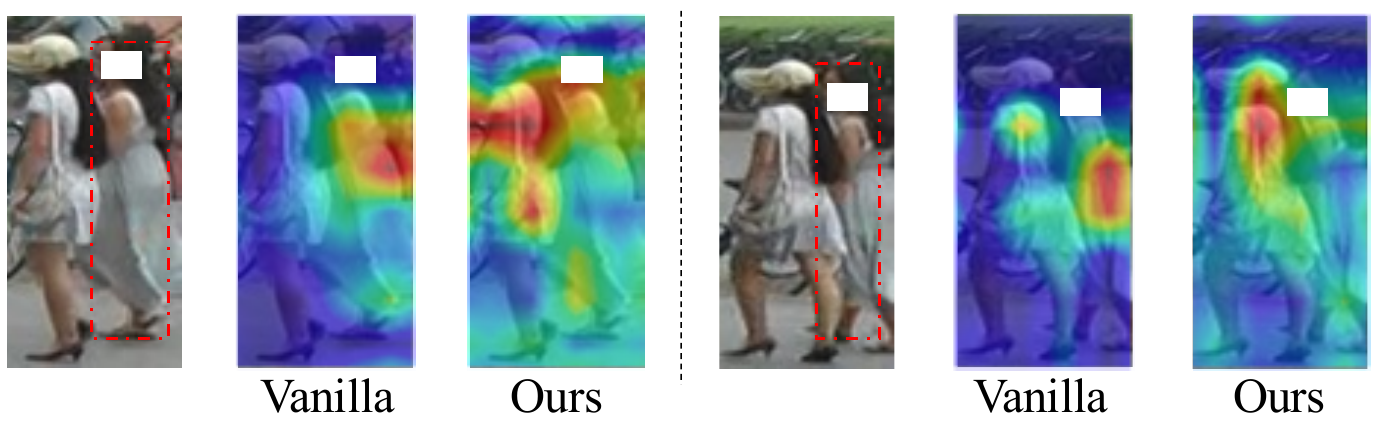}
    \caption{The attention maps of two images from the right figure of Fig.~\ref{fig:vis_tem}. These two images are notable for the vanilla Transformer in the tracklet because of distractors. }
   \label{fig:vis_exam}
\end{figure}

\subsubsection{Comparison on Spatial Attention}
As shown in Fig.~\ref{fig:vis_spa}, we draw the attention maps with the vanilla Transformer models and compare them with attention maps after adding the spatial constraints.
The attention weights are extracted from the attention matrix in the Spatial Transformer.
The first row of the matrix,~\emph{i.e.}, classification-token-related weights, is resized to reflect the importance of local patches.
The given examples are all successful detection bounding boxes, and the human bodies are occupying most of the images.
Despite this, the vanilla Transformer learns biased attention and discards many useful regions, indicating it suffers from serious over-fitting.
In comparison, with our proposed constrained spatial attention learning, we set up a more challenging objective for models to learn.
We encourage every image and every horizontal part can be classified correctly, so more regions on human bodies are covered by the attention maps.

\subsubsection{Comparison on Temporal Attention}
In Fig.~\ref{fig:vis_tem}, the left figure shows that in a tracklet where most of the images are of good quality, the vanilla Transformer without extra constraints tends to pay attention to very few frames.
For example, only the $6\mathrm{th}$ frame has a high response in the attention matrix of the Temporal Transformer.
Our proposed temporal attention loss makes the weights more evenly distributed across all frames via increasing the information entropy of attention weights.
We also set up an explicit upper bound of information entropy.
Therefore, there still exists space for the Transformer to distinguish less important frames, like the $8\mathrm{th}$ frame in the left example of Fig.~\ref{fig:vis_tem}.

Another benefit the temporal constraint brings is that the final representation of the tracklet should be shared by most frames.
In the right figure of Fig.~\ref{fig:vis_tem}, the vanilla Transformer gives high attention weight to the $6\mathrm{th}$ frame.
We observe the spatial attention map of this frame (as shown in Fig.~\ref{fig:vis_exam}) and find that the person in the red box is considered as the object of the tracklet.
From the whole tracklet, we know that the red box contains a distractor and should be ignored by the model.
With the temporal constraint, we require the output tokens of every frame within a tracklet to be as close as possible.
In this way, distractors are paid less attention because they only appear in some frames.
From Fig.~\ref{fig:vis_exam}, we see that with constrained attention learning, the model mainly extracts features from the target person.

\subsection{Comparison with Other Methods}

\begin{table}[t]
    \centering
    \setlength{\tabcolsep}{0.12cm}
    \begin{tabular}{l|cc|cc}
        \thickhline
    \multicolumn{1}{l|}{\multirow{2}[0]{*}{Methods}} & \multicolumn{2}{c|}{MARS} & \multicolumn{2}{c}{Duke} \\
    \cline{2-5}
          & rank-1 & mAP   & rank-1 & mAP \\
          \hline
            TKP~\cite{TKP}   & 84.0  & 73.3  & 94.0  & 91.7  \\
            STA~\cite{STA}   & 86.2  & 81.2  & 96.0  & 95.0  \\
            GLTR~\cite{GLTR} & 87.0 & 78.5 & 96.3 & 93.7  \\
            MG-RAFA~\cite{MGRAFA} & 88.8  & 85.9  & -     & - \\
            STE-NVAN~\cite{STENVAN} & 88.9  & 81.2  & 95.2  & 93.5  \\
            AGRL~\cite{AGRL}  & 89.5  & 81.9  & 97.0  & 95.4  \\
            NL-AP3D~\cite{NLAP3D} & 90.7  & 85.6  & 97.2  & 96.1  \\
            STGCN~\cite{STGCN} & 90.0  & 83.7  & 97.3  & 95.7  \\
            GRL~\cite{GRL}   & \textbf{91.0} & 84.8  & -     & - \\
            \hline
            Ours  & 88.7  & \textbf{86.3} & \textbf{97.6} & \textbf{97.4} \\
        \thickhline
    \end{tabular}%
  \caption{Comparison with recent works on MARS and Duke datasets.}
  \label{tab:compare1}%
\end{table}%

\begin{table}[t]
    \centering
    \setlength{\tabcolsep}{0.12cm}
    \begin{tabular}{l|cccc}
        \thickhline
        Methods & rank-1 & rank-5 & rank-10 & mAP \\
        \hline
        Two-stream~\cite{Simonyan2014TwoStreamCN} & 48.2  & 68.7  & 75.1  & 32.1  \\
        LSTM~\cite{Yan2016PersonRV} & 52.1  & 72.6  & 78.9  & 35.9  \\
        STMP~\cite{Liu2019SpatialAT}  & 56.8  & 76.2  & 82.0  & 39.1  \\
        M3D~\cite{Li2019Multiscale3C}   & 57.7  & 76.1  & 83.4  & 40.1  \\
        GLTR~\cite{GLTR}  & 63.1  & 77.2  & 83.8  & 44.3  \\
        \hline
        Ours  & \textbf{87.5} & \textbf{95.0} & \textbf{96.5} & \textbf{78.0} \\
        \thickhline
    \end{tabular}%
    \caption{Comparison with recent works on LS-VID.}
    \label{tab:compare2}%
\end{table}%

In Tab.~\ref{tab:compare1} and Tab.~\ref{tab:compare2}, we compare our method with recent video-based person ReID methods on the three benchmark datasets.
On both Duke and LS-VID, our proposed framework outperforms state-of-the-art methods in rank-1 accuracy and mAP.
For example, we surpass the graph-based method STGCN~\cite{STGCN} by $1.7\%$ in mAP on Duke.
This improvement is significant, considering the baseline accuracy of Duke is rather high.
Our method also outperforms GLTR~\cite{GLTR} by $24.4\%$ in rank-1 accuracy on LS-VID.
On MARS dataset, our method achieves state-of-the-art performance in mAP.
From other methods, we see that there is no absolutely positive correlation between rank-1 accuracy and mAP.
For instance, GRL~\cite{GRL} achieves the best rank-1 accuracy, but the mAP of GRL is significantly lower than MG-RAFA~\cite{MGRAFA}.
In MARS, of which detection boxes and tracking sequences are automatically generated by algorithms, the quality of images and tracklets may vary largely.
Therefore, different algorithms may emphasize different aspects.
Rank-1 accuracy reflects the ability to find the most confident positive sample, while mAP is a more comprehensive indicator, as it considers the ranking positions of all positive samples.
Therefore, the significant improvement in mAP on MARS validates the effectiveness of our proposed method.

\section{Conclusions}

This paper presents a simple yet effective Transformer-based representation learning framework for video-based person ReID tasks. 
In this framework, the constrained attention learning scheme and global attention learning branch are proposed to exploit spatial and temporal knowledge from ReID videos.
Furthermore, we introduce synthesized data pre-training for better initializing our framework and reducing the over-fitting risk.
Extensive experiments have shown the effectiveness of our proposed approaches in this paper. 
Importantly, to our best knowledge, it is the first work to evaluate the ability of the Transformer in video-based ReID tasks, which paves a new way for the application of the Transformer on more video-based tasks.

{\small
\bibliographystyle{ieee_fullname}
\bibliography{egbib}

\begin{thebibliography}{10}\itemsep=-1pt

\bibitem{GPT3}
T. Brown, B. Mann, Nick Ryder, Melanie Subbiah, Jared Kaplan, Prafulla
  Dhariwal, Arvind Neelakantan, Pranav Shyam, Girish Sastry, Amanda Askell,
  Sandhini Agarwal, Ariel Herbert-Voss, G. Kr{\"u}ger, T. Henighan, R. Child,
  Aditya Ramesh, Daniel~M. Ziegler, Jeffrey Wu, Clemens Winter, Christopher
  Hesse, Mark Chen, E. Sigler, Mateusz Litwin, Scott Gray, Benjamin Chess, J.
  Clark, Christopher Berner, Sam McCandlish, A. Radford, Ilya Sutskever, and
  Dario Amodei.
\newblock Language models are few-shot learners.
\newblock In {\em NeurIPS}, 2020.

\bibitem{SONA}
B. Bryan, Y. Gong, Y. Zhang, and C. Poellabauer.
\newblock Second-order non-local attention networks for person
  re-identification.
\newblock In {\em ICCV}, 2019.

\bibitem{DETR}
Nicolas Carion, Francisco Massa, Gabriel Synnaeve, Nicolas Usunier, Alexander
  Kirillov, and Sergey Zagoruyko.
\newblock End-to-end object detection with transformers.
\newblock {\em ArXiv}, abs/2005.12872, 2020.

\bibitem{autoaug}
E. Cubuk, Barret Zoph, Dandelion Man{\'e}, Vijay Vasudevan, and Quoc~V. Le.
\newblock Autoaugment: Learning augmentation strategies from data.
\newblock In {\em CVPR}, 2019.

\bibitem{videolstm}
J. Dai, Pingping Zhang, D. Wang, H. Lu, and H. Wang.
\newblock Video person re-identification by temporal residual learning.
\newblock {\em IEEE TIP}, 28:1366--1377, 2019.

\bibitem{GMMCP}
Afshin Dehghan, S.~M. Assari, and M. Shah.
\newblock Gmmcp tracker: Globally optimal generalized maximum multi clique
  problem for multiple object tracking.
\newblock In {\em CVPR}, 2015.

\bibitem{ImageNet}
Jia Deng, Wei Dong, Richard Socher, Li-Jia Li, Kai Li, and Li Fei-Fei.
\newblock Imagenet: A large-scale hierarchical image database.
\newblock In {\em CVPR}, 2009.

\bibitem{BERT}
J. Devlin, Ming-Wei Chang, Kenton Lee, and Kristina Toutanova.
\newblock Bert: Pre-training of deep bidirectional transformers for language
  understanding.
\newblock In {\em NAACL-HLT}, 2019.

\bibitem{ViT}
A. Dosovitskiy, Lucas Beyer, Alexander Kolesnikov, Dirk Weissenborn, Xiaohua
  Zhai, Thomas Unterthiner, M. Dehghani, Matthias Minderer, Georg Heigold, S.
  Gelly, Jakob Uszkoreit, and N. Houlsby.
\newblock An image is worth 16x16 words: Transformers for image recognition at
  scale.
\newblock In {\em ICLR}, 2021.

\bibitem{DPM}
Pedro~F. Felzenszwalb, Ross~B. Girshick, David~A. McAllester, and D. Ramanan.
\newblock Object detection with discriminatively trained part based models.
\newblock {\em IEEE TPAMI}, 32:1627--1645, 2009.

\bibitem{STA}
Y. Fu, Xiaoyang Wang, Yunchao Wei, and Thomas Huang.
\newblock Sta: Spatial-temporal attention for large-scale video-based person
  re-identification.
\newblock In {\em AAAI}, 2019.

\bibitem{actortrans}
Kirill Gavrilyuk, R. Sanford, M. Javan, and Cees G.~M. Snoek.
\newblock Actor-transformers for group activity recognition.
\newblock In {\em CVPR}, 2020.

\bibitem{NLAP3D}
Xinqian Gu, H. Chang, Bingpeng Ma, Hongkai Zhang, and X. Chen.
\newblock Appearance-preserving 3d convolution for video-based person
  re-identification.
\newblock In {\em ECCV}, 2020.

\bibitem{TKP}
Xinqian Gu, Bingpeng Ma, H. Chang, S. Shan, and X. Chen.
\newblock Temporal knowledge propagation for image-to-video person
  re-identification.
\newblock In {\em ICCV}, 2019.

\bibitem{resnet}
Kaiming He, Xiangyu Zhang, Shaoqing Ren, and Jian Sun.
\newblock Deep residual learning for image recognition.
\newblock In {\em CVPR}, 2016.

\bibitem{transreid}
Shuting He, Hao Luo, Pichao Wang, Fan Wang, Hao Li, and Wei Jiang.
\newblock Transreid: Transformer-based object re-identification, 2021.

\bibitem{InDefense}
Alexander Hermans, Lucas Beyer, and Bastian Leibe.
\newblock In defense of the triplet loss for person re-identification.
\newblock {\em arXiv preprint arXiv:1703.07737}, 2017.

\bibitem{ADAM}
Diederik~P Kingma and Jimmy Ba.
\newblock Adam: A method for stochastic optimization.
\newblock {\em arXiv preprint arXiv:1412.6980}, 2014.

\bibitem{GLTR}
Jianing Li, J. Wang, Q. Tian, Wen Gao, and S. Zhang.
\newblock Global-local temporal representations for video person
  re-identification.
\newblock In {\em ICCV}, 2019.

\bibitem{JVTC}
Jianing Li and Shiliang Zhang.
\newblock Joint visual and temporal consistency for unsupervised domain
  adaptive person re-identification.
\newblock In {\em ECCV}, 2020.

\bibitem{Li2019Multiscale3C}
Jianing Li, Shiliang Zhang, and Tiejun Huang.
\newblock Multi-scale 3d convolution network for video based person
  re-identification.
\newblock In {\em AAAI}, 2019.

\bibitem{Li2018DiversityRS}
Shuang Li, Slawomir Bak, P. Carr, and Xiaogang Wang.
\newblock Diversity regularized spatiotemporal attention for video-based person
  re-identification.
\newblock In {\em CVPR}, 2018.

\bibitem{STENVAN}
C. Liu, Chih-Wei Wu, Yu-Chiang~Frank Wang, and S. Chien.
\newblock Spatially and temporally efficient non-local attention network for
  video-based person re-identification.
\newblock In {\em BMVC}, 2019.

\bibitem{GRL}
Xuehu Liu, Pingping Zhang, Chenyang Yu, Huchuan Lu, and Xiaoyun Yang.
\newblock Watching you: Global-guided reciprocal learning for video-based
  person re-identification.
\newblock In {\em CVPR}, 2021.

\bibitem{Liu2019SpatialAT}
Yiheng Liu, Zhenxun Yuan, W. Zhou, and H. Li.
\newblock Spatial and temporal mutual promotion for video-based person
  re-identification.
\newblock In {\em AAAI}, 2019.

\bibitem{Loshchilov2019DecoupledWD}
I. Loshchilov and F. Hutter.
\newblock Decoupled weight decay regularization.
\newblock In {\em ICLR}, 2019.

\bibitem{RNN1}
N. McLaughlin, J. Rinc{\'o}n, and P. Miller.
\newblock Recurrent convolutional network for video-based person
  re-identification.
\newblock In {\em CVPR}, 2016.

\bibitem{fasterrcnn}
S. {Ren}, K. {He}, R. {Girshick}, and J. {Sun}.
\newblock Faster r-cnn: Towards real-time object detection with region proposal
  networks.
\newblock {\em IEEE TPAMI}, 39(6):1137--1149, 2017.

\bibitem{STGCN}
Jin rui Yang, W. Zheng, Q. Yang, Y. Chen, and Q. Tian.
\newblock Spatial-temporal graph convolutional network for video-based person
  re-identification.
\newblock In {\em CVPR}, 2020.

\bibitem{PSE}
M~Saquib Sarfraz, Arne Schumann, Andreas Eberle, and Rainer Stiefelhagen.
\newblock A pose-sensitive embedding for person re-identification with expanded
  cross neighborhood re-ranking.
\newblock In {\em CVPR}, 2018.

\bibitem{Simonyan2014TwoStreamCN}
K. Simonyan and Andrew Zisserman.
\newblock Two-stream convolutional networks for action recognition in videos.
\newblock In {\em NeurIPS}, 2014.

\bibitem{PDC}
Chi Su, Jianing Li, Shiliang Zhang, Junliang Xing, Wen Gao, and Qi Tian.
\newblock Pose-driven deep convolutional model for person re-identification.
\newblock In {\em ICCV}, 2017.

\bibitem{PCB}
Yifan Sun, Liang Zheng, Yi Yang, Qi Tian, and Shengjin Wang.
\newblock Beyond part models: Person retrieval with refined part pooling (and a
  strong convolutional baseline).
\newblock In {\em ECCV}, 2018.

\bibitem{DeiT}
Hugo Touvron, M. Cord, M. Douze, Francisco Massa, Alexandre Sablayrolles, and
  H. J{\'e}gou.
\newblock Training data-efficient image transformers {\&} distillation through
  attention.
\newblock {\em ArXiv}, abs/2012.12877, 2020.

\bibitem{aayn}
Ashish Vaswani, Noam Shazeer, Niki Parmar, Jakob Uszkoreit, Llion Jones,
  Aidan~N. Gomez, L. Kaiser, and Illia Polosukhin.
\newblock Attention is all you need.
\newblock In {\em NeurIPS}, 2017.

\bibitem{MGN}
Guanshuo Wang, Yufeng Yuan, Xiong Chen, Jiwei Li, and Xi Zhou.
\newblock Learning discriminative features with multiple granularities for
  person re-identification.
\newblock In {\em ACMMM}, 2018.

\bibitem{PTGAN}
Longhui Wei, Shiliang Zhang, Wen Gao, and Qi Tian.
\newblock Person transfer gan to bridge domain gap for person
  re-identification.
\newblock In {\em CVPR}, 2018.

\bibitem{GLAD}
Longhui Wei, Shiliang Zhang, Hantao Yao, Wen Gao, and Qi Tian.
\newblock Glad: global-local-alignment descriptor for pedestrian retrieval.
\newblock In {\em ACMMM}, 2017.

\bibitem{AGRL}
Yiming Wu, Omar El~Farouk Bourahla, X. Li, F. Wu, and Q. Tian.
\newblock Adaptive graph representation learning for video person
  re-identification.
\newblock {\em IEEE TIP}, 29:8821--8830, 2020.

\bibitem{dukevideo}
Yu Wu, Yutian Lin, Xuanyi Dong, Yan Yan, Wanli Ouyang, and Yi Yang.
\newblock Exploit the unknown gradually: One-shot video-based person
  re-identification by stepwise learning.
\newblock In {\em CVPR}, 2018.

\bibitem{RNN3}
Yichao Yan, B. Ni, Zhichao Song, Chao Ma, Yan Yan, and X. Yang.
\newblock Person re-identification via recurrent feature aggregation.
\newblock In {\em ECCV}, 2016.

\bibitem{Yan2016PersonRV}
Yichao Yan, B. Ni, Zhichao Song, Chao Ma, Yan Yan, and X. Yang.
\newblock Person re-identification via recurrent feature aggregation.
\newblock In {\em ECCV}, 2016.

\bibitem{MGH}
Yichao Yan, J. Qin, Jiaxin Chen, Li Liu, F. Zhu, Ying Tai, and Ling Shao.
\newblock Learning multi-granular hypergraphs for video-based person
  re-identification.
\newblock In {\em CVPR}, 2020.

\bibitem{TTSR}
Fuzhi Yang, Huan Yang, J. Fu, Hongtao Lu, and B. Guo.
\newblock Learning texture transformer network for image super-resolution.
\newblock In {\em CVPR}, 2020.

\bibitem{lidar3d}
Junbo Yin, J. Shen, Chenye Guan, Dingfu Zhou, and Ruigang Yang.
\newblock Lidar-based online 3d video object detection with graph-based message
  passing and spatiotemporal transformer attention.
\newblock In {\em CVPR}, 2020.

\bibitem{SCT}
Tianyu Zhang, Lingxi Xie, Longhui Wei, Yongfei Zhang, Bo Li, and Qi Tian.
\newblock Single camera training for person re-identification.
\newblock In {\em AAAI}, 2020.

\bibitem{UnrealPerson}
Tianyu Zhang, Lingxi Xie, Longhui Wei, Zijie Zhuang, Yongfei Zhang, Bo Li, and
  Qi Tian.
\newblock Unrealperson: An adaptive pipeline towards costless person
  re-identification.
\newblock In {\em CVPR}, 2021.

\bibitem{MGRAFA}
Zhizheng Zhang, Cuiling Lan, Wenjun Zeng, and Zhibo Chen.
\newblock Multi-granularity reference-aided attentive feature aggregation for
  video-based person re-identification.
\newblock In {\em CVPR}, 2020.

\bibitem{MARS}
L. Zheng, Zhi Bie, Y. Sun, Jingdong Wang, Chi Su, S. Wang, and Q. Tian.
\newblock Mars: A video benchmark for large-scale person re-identification.
\newblock In {\em ECCV}, 2016.

\bibitem{MARKET}
Liang Zheng, Liyue Shen, Lu Tian, Shengjin Wang, Jingdong Wang, and Qi Tian.
\newblock Scalable person re-identification: A benchmark.
\newblock In {\em ICCV}, 2015.

\bibitem{RandomErasing}
Z. Zhong, L. Zheng, Guoliang Kang, Shaozi Li, and Y. Yang.
\newblock Random erasing data augmentation.
\newblock In {\em AAAI}, 2020.

\bibitem{RPT2}
K. Zhou, Yongxin Yang, A. Cavallaro, and T. Xiang.
\newblock Learning generalisable omni-scale representations for person
  re-identification.
\newblock {\em ArXiv}, abs/1910.06827, 2019.

\bibitem{RPT1}
K. Zhou, Yongxin Yang, A. Cavallaro, and T. Xiang.
\newblock Omni-scale feature learning for person re-identification.
\newblock In {\em ICCV}, 2019.

\bibitem{RNN2}
Z. Zhou, Y. Huang, Wei Wang, Liang Wang, and T. Tan.
\newblock See the forest for the trees: Joint spatial and temporal recurrent
  neural networks for video-based person re-identification.
\newblock In {\em CVPR}, 2017.

\bibitem{CBN}
Zijie Zhuang, Longhui Wei, Lingxi Xie, Tianyu Zhang, Hengheng Zhang, Haozhe Wu,
  Haizhou Ai, and Qi Tian.
\newblock Rethinking the distribution gap of person re-identification with
  camera-based batch normalization.
\newblock In {\em ECCV}, 2020.

\end{thebibliography}
}

\end{document}